# Rotational Mutation Genetic Algorithm on optimization Problems

Masoumeh Vali

Department of Mathematics, Dolatabad Branch, Islamic Azad University, Isfahan, Iran
E-mail: vali.masoumeh@gmail.com

**Abstract**

Optimization problem, nowadays, have more application in all major but they have problem in computation. Calculation of the optimum point in the spaces with the above dimensions is very time consuming. In this paper, there is presented a new approach for the optimization of continuous functions with rotational mutation that is called RM. The proposed algorithm starts from the point which has best fitness value by elitism mechanism. Then, method of rotational mutation is used to reach optimal point. In this paper, RM algorithm is implemented by GA(Briefly RMGA) and is compared with other well- known algorithms: DE, PGA, Grefensstette and Eshelman [15, 16] and numerical and simulation results show that RMGA achieve global optimal point with more decision by smaller generations.

1. Introduction

In practical projects, we always try to find the optimal solution. As compared to other optimization methods, Genetic Algorithm (GA) as an auto-adapted global searching system by simulating biological evolution and the fittest principle in natural environment seeks best solution more effectively. Therefore, GA is a heuristic search technique that mimics the natural evolution process such as selection, crossover and mutation operations. The selection pressure drives the population toward better solutions while crossover uses genes of selected parents to produce offspring that will form the next generation. Mutation is used for avoiding of premature convergence and consequently escaping from local optimal. The GAs has been very successful in handling combinatorial optimization problems which are difficult [1,2].

There is a lot of attention to find optimization problems to the success of new algorithms for solving large classes of problems from diverse areas such as structural optimization, computer sciences, operations research, economics, and engineering design and control.

This paper concerned with the following simple-bounded continuous optimization problem:
f ($x_1, x_2, ..., x_m$) where each $x_i$ is a real parameter so that $a_i \leq x_i \leq b_i$ for some constants $a_i$ and $b_i$. This problem has widespread applications including optimization simulating models, fitting nonlinear curve to data, solving system of nonlinear, engineering design and control problem, and setting weights on neural networks.

RM is an approach for finding global optimal point (max/min) by simple algebra, without derivation and based on search and for more optimal performance, RM is implemented by GA and resulting algorithm is performed as follows:

- Finding vertex of polytope has best fitness.
- Global optimal point is obtained by using rotational mutation operator.

This paper starts with the description of related work in section 2. Section 3 gives the outline of Model and Problem Definition RM. In section 4, test problems of RM method is implemented. In section 5, Schemata Analysis for RMGA is present. Evaluation by De Jong's functions and the compression of RMGA with the other methods (DE, PGA, Grefensstette and Eshelman [8, 9]) for De Jong Functions are shown in sections 6. The discussion ends with a conclusion and future trend.

2. **Related Work**

Mutation and the mutation probability ($P_m$) are important parameters in GAs. The mutation operator generates a new string by altering one or more bits of a string. By applying the mutation operator to a string, muting each bit of the string independently from the other bits is considered. So, the mutation operator is more likely to significantly disrupt the allocation of trials to high order schemata than to low order ones. The efficiency of the mutation operator as a means of exploring the search space is questionable. A GA using mutation as the only genetic operator would be a random search that is biased toward sampling good hyper planes rather than poor ones [3].
Mutation is a genetic operator that alters one or more gene values in a chromosome from its initial state. This can result in entirely new gene values being added to the gene pool. With these new gene values, the genetic algorithm may be able to arrive at better solution than was previously possible. Mutation is an important part of the genetic search as help helps to prevent the population from stagnating at any local optima.

Pratibha Bajpal and Manojkumar [4] have proposed an approach to solve Global Optimization Problems by GA and obtained that GA is applicable to both continuous and discrete optimization problems.

Hayes and Gedeon [5] considered infinite population model for GA where the generation of the algorithm corresponds to a generation of a map. They showed that for a typical mixing operator all the fixed points are hyperbolic.

Gedeon et al. [6] showed that for an arbitrary selection mechanism and a typical mixing operator, their composition has finitely many fixed points.

Qian et al.[7] proposed a GA to treat with such constrained integer programming problem for the sake of efficiency. Then, the fixed-point evolved (E)-UTRA PRACH detector was presented, which further underlines the feasibility and convenience of applying this methodology to practice.

Devis Karaboga and Selcuk.[8] is proposed new heuristic approach with deferential evaluation (DE) for finding a true global minimum regardless of the initial parameter values , fast convergence using similar operators crossover, mutation and selection.

### 3. Model and Problem Definition RM

Consider f($x_1, x_2, x_3, \ldots, x_n$) with constraint $a_i \leq x_i \leq b_i$ for i=1, 2,…, n. The serial algorithm RM is as follow:

**Step 1:** Draw the diagrams for $x_i = b_i$ and $x_i = a_i$ for i=1, 2,…, n.

**Step2:** Consider the vertex of this polytope which has best fitness among other vertexes and call it S.

**Step3:** Put S′ = f(S) , $\vec{r_0}$=(1,1,…,1), (1,-1,1,…,1),…, or (-1,-1,…,-1) and
$r_n = \alpha\vec{r_0}$, = 0.1n , n=1,2,…,10.

**Step 4:** Move the point 'S' with length of α in direction of $\vec{r_0}$ vector (notice: the direction of $\vec{r_0}$ vector must be inside the search space, also the α measurement depends on problem precision). The endpoint of vector $r_n$ is called P.

**Step4:** If f (P) better than S′ , then S=P and go to step 7 else go to step 5.

**Step 5:** Put $\beta = 0.1, 0.25, \ldots, e_n = (1,1,\ldots,1), (1,-1,1,\ldots,1), \ldots,$ and $(-1,-1,\ldots,-1)$, P = $\beta e_n$.

**Step 6:** If f (P) is better than S′ , then S=P else go to step 7.

**Step7:** If P is in search space, go to step 2 else go to stop.

## 4. Implementing Test Problems by RM

### 4.1. Test problem1:

Consider Beale Function as follow:

$$f(x) = (1.5 - x_1 + x_1 x_2)^2 + (2.26 - x_1 + x_1 x_2^2)^2 + (2.625 - x_1 + x_1 x_2^3)^2,$$

such that $-4.5 \leq x_i \leq 4.5$, $i = 1, 2$. The global minimum: $x^* = (3, 0.5)$, $f(x^*) = 0$. Beal function's graph is as follow.

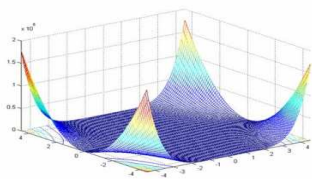

Figure1: Beale Function

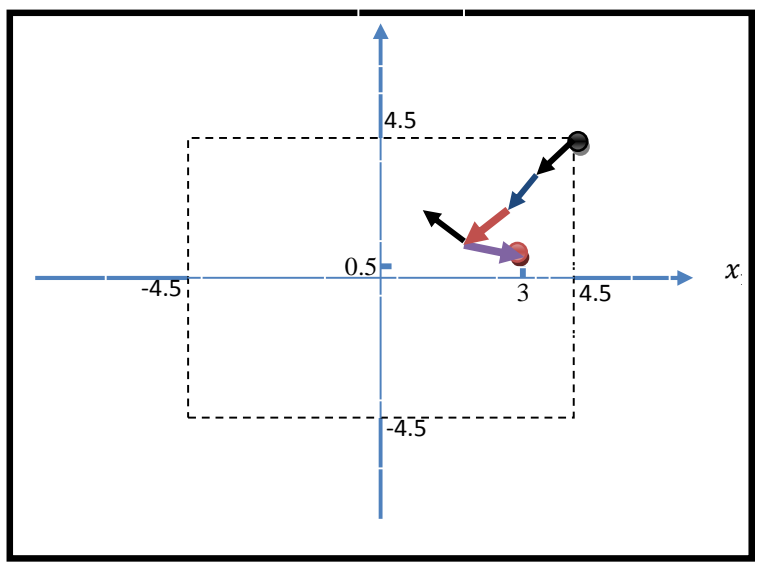

Figure 2: The Performance RM

As you can see in Figure 2, algorithm starts from point (4.5, 4.5) which has the best fitness among other points: (-4.5, 4.5), (-4.5, -4.5), (4.5, -4.5), after that, by using rotational mutation the global optimal point is funded.

### 1.1. Test problem2:

This function is a continuous and uni-modal function. The optimization problem is as follows:

$$\min f(x_1, x_2) = x_1{}^2 + (x_2 - 0.4)^2 \qquad\qquad -2 < x_i < 2, \quad i = 1,2$$

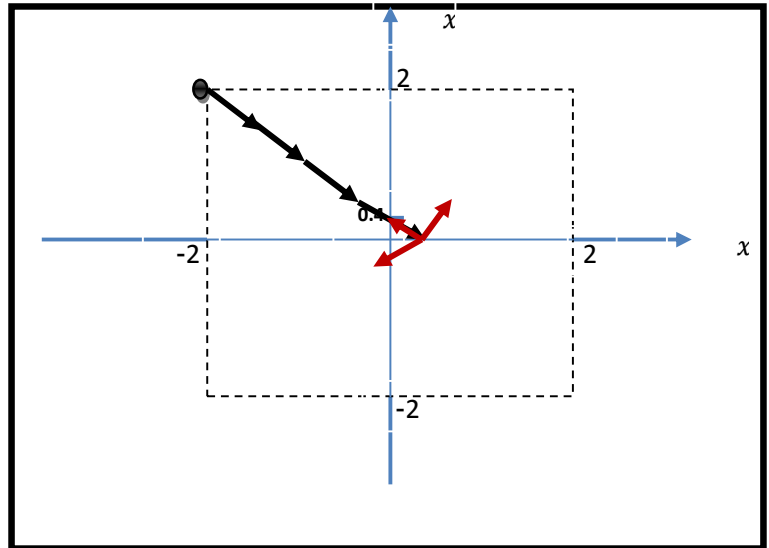

Figure 3: The Performance RMC

In Figure 3, as you can see, algorithm starts from the point (-2, 2) which has the best fitness than the rest of the points and after that by using rotational mutation the global optimal point is funded.

## 2. Schema Analysis for RMGA

This schema – as you can see in Figure 4- starts from the offspring (vertex) which has best fitness value that is called 'S'. Then it sets initial point of $\vec{r_0}$ vector at 'S' and mutates offspring 'S' in direction of $\vec{r_0}$ vector with length of mutation α.
(Notice: the direction of $\vec{r_0}$ vector must be inside the search space, also the α measurement depends on problem precision). This mutated offspring is called P.
If fitness value of offspring P was better than fitness value of offspring S we would use mutation operator for P point. Otherwise by using rotational mutation by $\vec{e_n}$ vector, we would search an Offspring-with better fitness value in comparison with 'S'. We repeat this action while the mutated offspring doesn't get out of search space. Eventually, we would select the last new produced offspring- inside the search space- as the global optimization point which would be fixed point of our question.

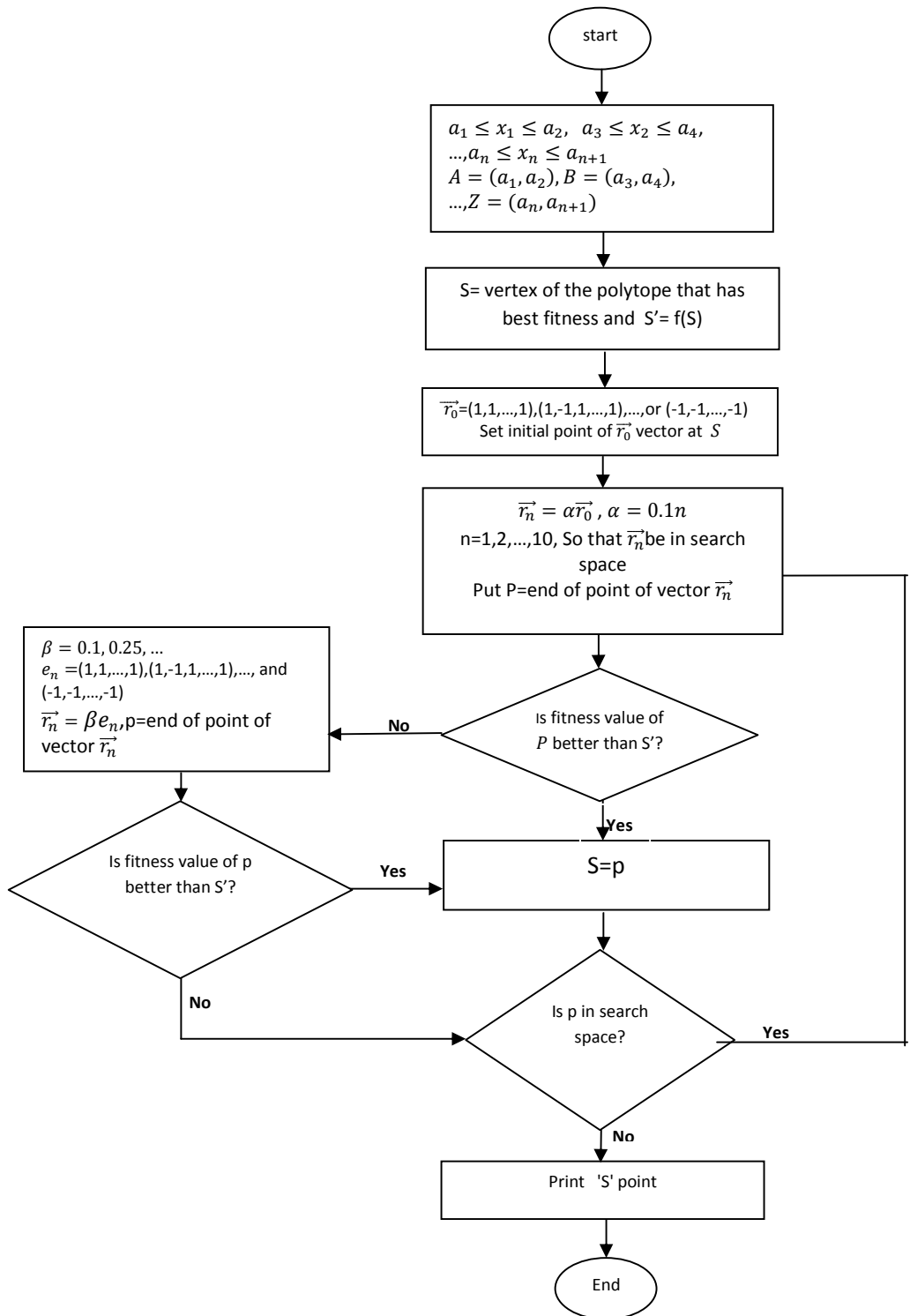

Figure 4: The Improved GA

## 3. Evaluation

In this section, Definition of De Jong's Functions, Numerical results of De Jongs' functions by RMGA and the compression of RMGA with the other methods (DE, PGA, Grefensstette and Eshelman [8, 9]) for De Jong Functions are presented.

### 6.1 De Jong's Functions

In this section, Definition of De Jong's Functions (F1 to F5) and initial population RMGA which depends on the dimensional space are shown in Table 1.

Table 1: De Jong's Functions

| Function Number | Function | Limits | Dim. | Initial Population |
|---|---|---|---|---|
| F1 | $\sum_{i=1}^{3} x_i^2$ | $-5.12 \leq x_i \leq 5.12$ | 3 | 8 |
| F2 | $100.(x_1^2 - x_2) + (1 - x_1)^2$ | $-2.048 \leq x_i \leq 2.048$ | 2 | 4 |
| F3 | $30. + \sum_{i=1}^{5} \lfloor x_j \rfloor$ | $-5.12 \leq x_i \leq 5.12$ | 5 | 32 |
| F4 | $\sum_{i=1}^{30} (ix_i^4. + Gauss\,(0,1))$ | $-1.28 \leq x_i \leq 1.28$ | 30 | |
| F5 | $\dfrac{1}{0.002 + \sum_{i=0}^{24} \dfrac{1}{i + \sum_{j=0}^{1}(x_j - a_{ij})^6}}$ | $-65.536 \leq x_i \leq 65.536$ | 2 | 4 |

### 6.2. Numerical Results

In this section, the experimental results of RMGA on the five problems of De Jong (F1 to F5) [De Jong, 1975] in Table 2 are shown. Furthermore, results were saved for the best performance (BP) which BP is the best fitness of the objective function obtained over all function evaluations. At last, standard deviation (SD) is calculated and measured with final answer of De Jong function. The following parameters were evaluated in the following table.
rotational mutation size (RMS), number of iterations rotational mutation (TRM).

Table 2: Numerical result of De Jongs' function by RMGA

| Algorithm | De Jong's Function | RMS | TRM | Best Point | BP | SD |
|---|---|---|---|---|---|---|
| RMGA | F1 | 0.1 | 78 | (0,0,0) | 0 | 0 |
| RMGA | F2 | 0.1 | 16 | (1,1) | 0 | 0 |
| RMGA | F3 | 0.1 | 7 | (-5.12,-5.12,-5.12) | 0 | 0 |
| RMGA | F4 | 0.1 | 195 | (0,0,…,0) 30 times | Depend on $\eta$ | 0 |
| RMGA | F5 | 0.1 | 340 | (-32,-32) | | 0 |

## 6.3. Simulating Results

Dervis Karaboga and Selcuk Ökdem[8] introduced new method DE for finding global optimization problems. They compared DE method with other methods: PGA, Grefensstette and Eshelman [9] and showed that DE algorithm works much better than other methods [9].
In order to obtain the average results, PGA, Grefensstette and Eshelman algorithms were run 50 times; the DE algorithm was run 1000 times and RMGA 350 times for each function.

Table 3: The average number of generations.

| Algorithms | F1 | F2 | F3 | F4 | F5 |
|---|---|---|---|---|---|
| $PGA(\lambda = 4)$ | 1170 | 1235 | 3481 | 3194 | 1256 |
| $PGA(\lambda = 8)$ | 1526 | 1671 | 3634 | 5243 | 2076 |
| Grefensstette | 2210 | 14229 | 2259 | 3070 | 4334 |
| Eshelman | 1538 | 9477 | 1740 | 4137 | 3004 |
| DE(F: RandomValues) | 260 | 670 | 125 | 2300 | 1200 |
| RMGA | 78 | 16 | 7 | 195 | 340 |
| PNG | 3.333 | 41.875 | 17.857 | 11.794 | 3.529 |

As you can see in Table 3, the convergence speed RMGA achieves the optimal point with more decision by smaller generation. Furthermore, the most significant improvement is with F2 since the proportion of the number of generations (PNG) about $\frac{670}{16} \cong 64$ times smaller than the average of the number of generations DE algorithm.

**Conclusion**

RMGA is a new method for finding the true optimal global optimization is based on rotational mutation (RM). In this work, the performance of the RMGA has been compared to that of some other well known GAs. From the simulation studies, it was observed that RMGA achieve the optimal point with more decision by smaller generation. Therefore, RMGA seems to be a promising approach for engineering optimization problems.